\documentclass[10pt,journal,compsoc]{IEEEtran}
\ifCLASSOPTIONcompsoc
  \usepackage[nocompress]{cite}
\else
  \usepackage{cite}
\fi
\ifCLASSINFOpdf
  \usepackage[pdftex]{graphicx}
  \usepackage{caption}
  \usepackage{amssymb,amsmath}
  \usepackage{subcaption}

  \usepackage[textsize=tiny]{todonotes}
  \usepackage[linesnumbered,ruled,vlined]{algorithm2e}
  \usepackage{algpseudocode} 

\usepackage{multicol}
\usepackage{multirow}
\usepackage{booktabs}
\usepackage{hyperref}
\else
  \usepackage[dvips]{graphicx}
  \usepackage[textsize=tiny]{todonotes}
  \usepackage{caption}
  \usepackage{amssymb,amsmath}
  \usepackage{subcaption}
  \usepackage{multicol}
\usepackage{multirow}
\usepackage{booktabs}
\usepackage[textsize=tiny]{todonotes}

\fi
\begin{document}
%
\title{
E2GNN: Efficient Graph Neural Network Ensembles for Semi-Supervised Classification \\
}
%
%
%

\author{Xin~Zhang,
        Daochen~Zha,
        Qiaoyu~Tan
\IEEEcompsocitemizethanks{
\IEEEcompsocthanksitem Xin Zhang is with the Department of Computing, The Hong Kong Polytechnic University, Hong Kong SAR.\protect\\
E-mail: see xin12.zhang@connect.polyu.hk


\IEEEcompsocthanksitem Daochen Zha is with the Department of Computer Science, Rice University, Texas, USA \protect\\
E-mail: see daochen.zha@rice.edu

\IEEEcompsocthanksitem Qiaoyu Tan is with Computer Science Department, New York University Shanghai, Shanghai.\protect\\
E-mail: see qiaoyu.tan@nyu.edu}
\thanks{Manuscript received April 19, 2005; revised August 26, 2015.}}

%
%

\markboth{Journal of \LaTeX\ Class Files,~Vol.~14, No.~8, August~2015}%
{Shell \MakeLowercase{\textit{et al.}}: Bare Demo of IEEEtran.cls for Computer Society Journals}
%



\IEEEtitleabstractindextext{%
\begin{abstract}
This work studies \textit{ensemble learning} for graph neural networks (GNNs) under the popular semi-supervised setting. Ensemble learning has shown superiority in improving the accuracy and robustness of traditional machine learning by combing the outputs of multiple weak learners.
However, adopting a similar idea to integrate different GNN models is challenging because of two reasons. First, GNN is notorious for its poor inference ability, so naively assembling multiple GNN models would deteriorate the inference efficiency. Second, when GNN models are trained with few labeled nodes, their performance are limited. In this case, the vanilla ensemble approach, e.g., majority vote, may be sub-optimal since most base models, i.e., GNNs, may make the wrong predictions. To this end, in this paper, we propose an efficient ensemble learner--E2GNN to assemble multiple GNNs in a learnable way by leveraging both labeled and unlabeled nodes. Specifically, we first pre-train different GNN models on a given data scenario according to the labeled nodes. Next, instead of directly combing their outputs for label inference, we train a simple multi-layer perceptron--MLP model to mimic their predictions on both labeled and unlabeled nodes. Then the unified MLP model is deployed to infer labels for unlabeled or new nodes. Since the predictions of unlabeled nodes from different GNN models may be incorrect, we develop a reinforced discriminator to effectively filter out those wrongly predicted nodes to boost the performance of MLP. By doing this, we suggest a principled approach to tackle the inference issues of GNN ensembles and maintain the merit of ensemble learning: improved performance. Comprehensive experiments over both transductive and inductive settings, across different GNN backbones and 8 benchmark datasets, demonstrate the superiority of E2GNN. Notably, our E2GNN shows good robustness towards feature- and topology-level perturbations. The code is anonymously released in \url{https://github.com/qiaoyu-tan/E2GNN}
\end{abstract}

}

\maketitle

\IEEEdisplaynontitleabstractindextext

%
\IEEEpeerreviewmaketitle

\IEEEraisesectionheading{\section{Introduction}\label{sec:introduction}}

\IEEEPARstart{G}{raphs} have been widely used to model a variety of structured and relational systems, ranging from social networks~\cite{zafarani2014social} to citation networks~\cite{namata2012query}, to molecular graphs~\cite{pei2020geom}. Among various analytical tasks in graph-structure data, node classification is an essential one that has a broad spectrum of applications, such as social circle learning~\cite{leskovec2012learning}, protein classification~\cite{borgwardt2005protein}, and document categorization~\cite{tang2008arnetminer}. 
Recently, graph neural networks (GNNs)~\cite{kipf2016semi,velivckovic2017graph,wu2019simplifying,chiang2019cluster,liu2020towards} have received significant attention for node classification due to their effectiveness in leveraging both labeled and unlabeled nodes. Numerous GNN variants~\cite{kipf2016semi,klicpera2018predict,velivckovic2017graph,wu2019simplifying,liu2020towards,dwivedi2021graph,chen2019powerful,xu2018powerful,hamilton2017inductive,tan2023collaborative} have been proposed in the past few years; these methods mainly differ in their message passing mechanisms~\cite{gilmer2017neural}. To name a few, GCN~\cite{kipf2016semi} defines message propagation as spectral filters on graphs~\cite{hammond2011wavelets}. GAT~\cite{velivckovic2017graph} adopts self-attention networks~\cite{vaswani2017attention} to achieve an adaptive propagation in a learnable way. APPNP~\cite{klicpera2018predict} derives an improved propagation schema based on personalized PageRank~\cite{page1999pagerank}.

\begin{figure}[t]
\centering
\includegraphics[width=7cm]{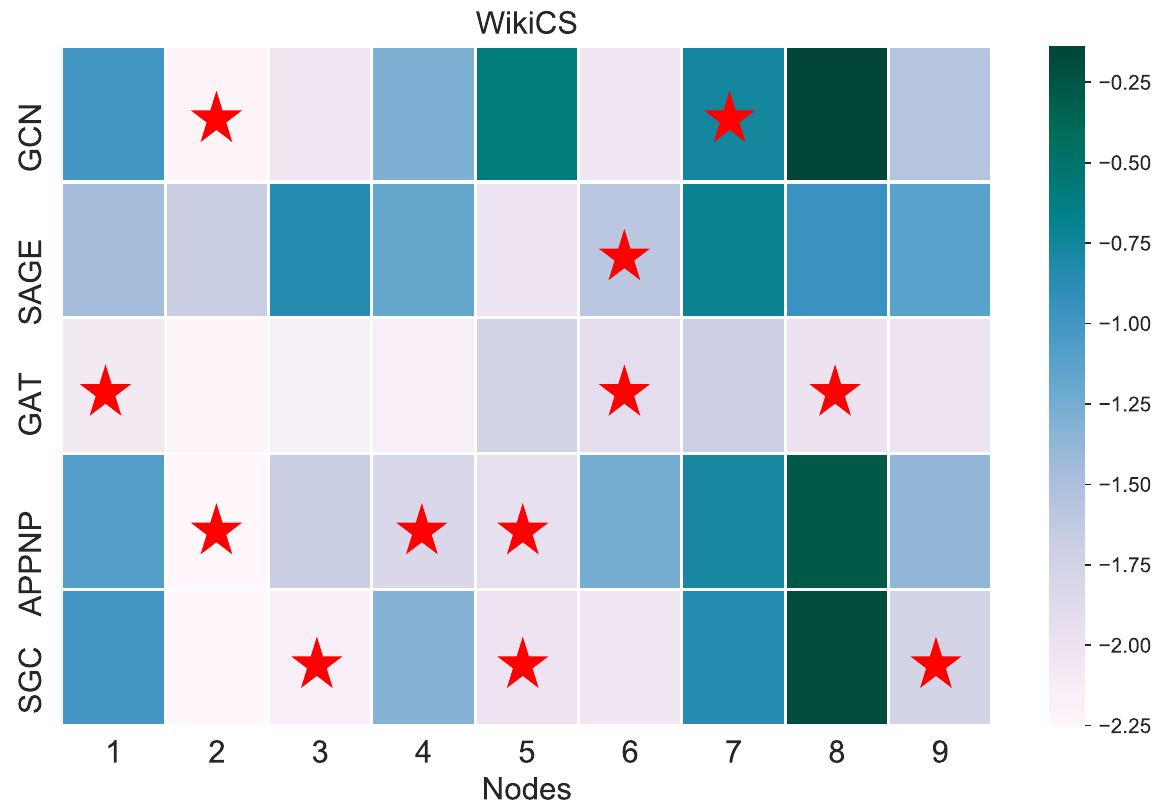}
\caption{The performance of five popular GNN models including GCN~\cite{kipf2016semi}, GAT~\cite{velivckovic2017graph}, APPNP~\cite{klicpera2018predict}, GraphSAGE~\cite{hamilton2017inductive}, and SGC~\cite{wu2019simplifying} on 9 random nodes. The color represents certainty, i.e., negative entropy; warmer colors indicate higher certainty. \textit{Red starts} (\textcolor{red}{$\bigstar$}) mark the correct prediction. GNNs tend to make wrong predictions with high certainty.}
\label{fig_toy}
\end{figure}
Despite their fruitful successes, no single proposal performs consistently well~\cite{sen2008collective,hu2021ogb} since real-world graphs are complex and each model could only capture a specific property of them. 
As a result, in practice, different GNN models may have different prediction abilities on different nodes. To verify this point, we conduct a preliminary experiment on five state-of-the-art GNN models and report their results in Figure~\ref{fig_toy}. We can observe that different models tend to make correct predictions (masked as \textcolor{red}{$\bigstar$}) on different nodes. For example, GCN can make accurate predictions on nodes 2 and 7, while GAT can correctly predict the labels of other nodes, i.e., nodes 1, 6, and 8. These results indicate the complementary nature of five GNN models sample-wisely. Therefore, if we can select a suitable model for each node to make the correct prediction, the overall accuracy results could be largely improved. This observation motivates us to integrate the predictions of different GNN models to boost their performance. 

In parallel, ensemble learning~\cite{dietterich2002ensemble,zhou2012ensemble} has been widely adopted in traditional machine learning to improve the performance of base classifiers. Typical examples include Random Forest~\cite{breiman2001random}, XGBoost~\cite{chen2015xgboost} and Generalized Boosted Models (GBM~\cite{ridgeway2007generalized}), in which the core idea is to combine the outputs of multiple individual tree classifiers to make the final prediction. Despite the simplicity, these methods have shown great success in a variety of practical scenarios~\cite{yang2021survey}, such as intrusion detection~\cite{dhaliwal2018effective} and healthcare~\cite{wang2020prediction}. Inspired by the effectiveness of ensemble learning and the diversity of GNN models, the following question naturally arises:

\textbf{\textit{Can we ensemble different GNN models to improve the performance and robustness of GNNs?}}

However, it is a non-trivial and challenging task to leverage different GNN models mainly because of two reasons: \textbf{(i) \textit{the poor inference ability of GNN.}} In general, GNN is notorious for its poor inference ability, especially when going deeper~\cite{chen2020simple} and applying it to large-scale graphs~\cite{hu2020open}. The primary reason is that the message propagation among neighboring nodes from multi-hops away incurs heavy data dependency~\cite{zhang2021graph}, leading to substantially computational costs and memory footprints. Therefore, naively assembling multiple GNN models would deteriorate the inference problems, hindering their use in resource-constrained applications, such as online serving~\cite{crankshaw2017clipper} and edge devices~\cite{baller2021deepedgebench}; \textbf{(ii) \textit{the limited performance of GNN models without sufficient labeled data.}} Since annotating nodes in a graph is expensive and time-consuming, GNN is often trained with few labeled nodes in real-world scenarios, leading to limited classification results~\cite{garcia2017few}. As a consequence, directly combing the predictions of different GNN models might be sub-optimal, since most of them may make the wrong predictions simultaneously, as shown in Figure~\ref{fig_motivation}. 
Therefore, a more effective ensemble approach is urgently needed to integrate GNN models under the semi-supervised setting.

To address the aforementioned challenges, we propose a novel ensemble learning framework -- E2GNN, which performs effective and efficient GNN ensemble via knowledge distillation. The high-level idea is to compress cumbersome GNN models into simple multi-layer perceptrons (MLPs) to accelerate the inference. To effectively transfer the knowledge of multiple GNNs to a unified student, we develop an agent-guided distillation schema via reinforcement learning. The key innovation lies in allowing the agent to adaptively \textit{select the best GNN model or reject all GNNs for distillation in a node-wise manner}. Throughout the training, the agent gradually learns to choose the most suitable GNN teacher for each node, and the student model learns to ensemble the knowledge (i.e., soft labels) of the teacher models under the agent's guidance. Our main \textbf{contributions} are as follows:
\begin{itemize}
    \item We study the ensemble learning problem of GNNs with limited labeled data, and propose a novel ensemble learner -- E2GNN. To the best of our knowledge, E2GNN is the first GNN ensemble work that achieves MLPs-level inference speed while maintaining the merits of ensemble learning.  
    \item E2GNN innovates to selectively utilize the soft labels of unlabeled nodes from different GNN models via a reinforced agent. By choosing the correctly predicted nodes and filtering out those incorrectly predicted by all GNN models, E2GNN can learn to mimic their best, leading to better results.
    \item Extensive experiments on several benchmark datasets across different GNN backbones demonstrate the superiority of our proposal. Notably, E2GNN not only outperforms state-of-the-art baselines in both transductive and inductive scenarios, but also shows good robustness w.r.t. graph noises, such as feature masking and edge perturbation.  
\end{itemize}

 

\section{Related Work}
In this section, we briefly review some related works in GNN ensemble learning and GNN compression and refer readers to~\cite{chen2020graph,wu2020comprehensive,xiao2022graph} for a comprehensive review of recent advances in GNNs.

\noindent\textbf{GNN Ensemble.} Recently, some studies~\cite{kosasih2021graph,nagarajan2022efficient,allen2020towards} have been made to integrate different GNN models via ensemble learning~\cite{dietterich2002ensemble}. For instance, GNNE~\cite{kosasih2021graph} suggests directly combining the outputs of multiple GNN models for node classification. GEENI~\cite{nagarajan2022efficient} proposes to improve the efficiency of GNNE by pruning both edges in the graph topology and model weights during the training. While GEENI could reduce the computational costs of GNN ensembles to some extent, the overall inference cost is still significantly higher than the standard GNN model.

\noindent\textbf{GNN Distillation.} Some efforts~\cite{yang2020distilling,yan2020tinygnn,deng2021graph,xu2020graphsail,tan2022double} have been devoted to improving the efficiency of GNNs via knowledge distillation~\cite{hinton2015distilling}. However, most of them are designed to distill large GNNs to smaller ones~\cite{yang2020distilling,yan2020tinygnn,feng2022freekd}, so the inference speedup is rather limited since the data dependency issue remains unresolved. To tackle this issue, GLNN~\cite{zhang2021graph} proposes to conduct cross-model knowledge distillation from the GNN model to simple multi-layer perceptrons (MLPs). Since MLPs are inference-friendly in modern systems, GLNN successfully extends the applicability of GNNs to real-world applications. However, the major limitation of GLNN is that it focuses on only one GNN model and cannot be directly applied to distill knowledge from multiple GNN models.

\begin{figure*}[t]
\begin{center}
\includegraphics[width=17cm, height=8cm]{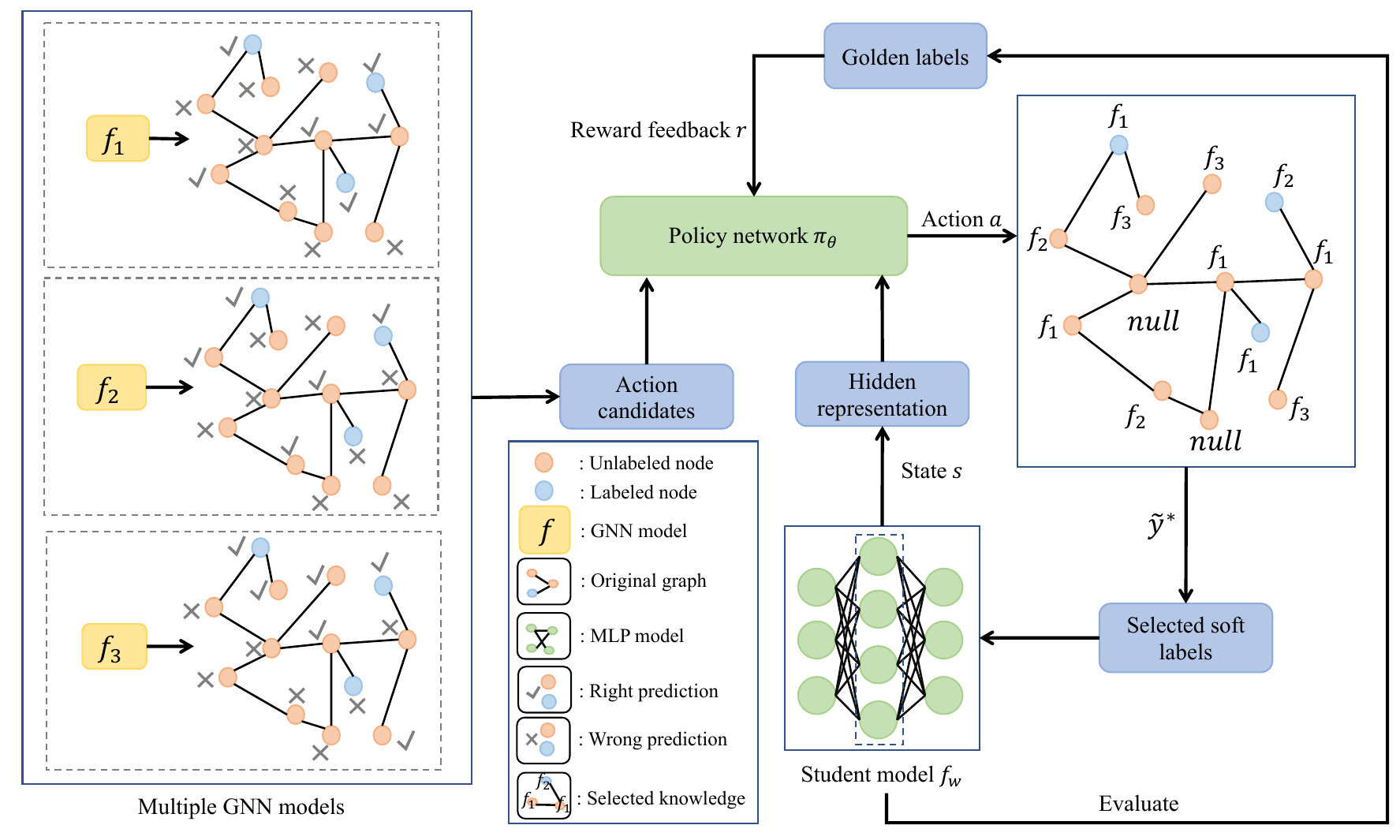}
\caption{The flowchart of the proposed E2GNN framework. Instead of averaging the output of GNN models, E2GNN compresses multiple GNN models into an inference-friendly MLP model via a node-level agent. The agent enables us to adaptively choose the correctly predicted nodes and filter out those incorrectly predicted by all GNN models (i.e., null action) for distillation.  }
\label{fig:e2gnn}
\end{center}
\end{figure*}

\section{Methodology}
In this section, we illustrate the details of our E2GNN framework, as shown in Figure~\ref{fig:e2gnn}. Before going into the details, we first provide the notations and preliminaries.

\subsection{Preliminaries}
\textbf{Notations.} Given a graph $\mathcal{G}=(\mathcal{V},\mathcal{E})$, where $\mathcal{V}$ is the node set and $\mathcal{E}$ is the edge set. Let $\mathbf{X}\in\mathbb{R}^{N\times F}$ be the feature matrix where each row $\mathbf{x}_v$ denotes the $F$-dimensional feature vector of node $v\in\mathcal{V}$. $\mathbf{Y}\in\mathbb{R}^{N\times C}$ is the label matrix for $N$ nodes, where $C$ is the number of distinct labels. $\mathbf{y}_v\in\mathbb{R}^C$ represents the one-hot label vector of node $v$. Without losing generality, among $N$ nodes, we assume only a small portion of nodes are labeled while the majority of the nodes are unlabeled. Throughout this paper, we mark labeled nodes with superscript $L$, i.e., $\mathcal{V}^L$, $\mathbf{X}^L$, and $\mathbf{Y}^L$, and unlabeled nodes with superscript $U$, i.e., $\mathcal{V}^U$, $\mathbf{X}^U$, and $\mathbf{Y}^U$. We use $f(\cdot)$ to denote an arbitrary GNN model. In semi-supervised setting, $f(\cdot)$ will be trained according to the labeled nodes, i.e., $\mathbf{Y}^L$.

In our ensemble learning scenario, we are given $K$ pre-trained GNN models $\{f_k(\cdot)\}_{k=1}^K$. The output (i.e., soft labels) of the $k$-th GNN model for a given node $v$ is written as $\widetilde{\mathbf{y}}_v^k=[\widetilde{\mathbf{y}}_{v,1}^k,\cdots,\widetilde{\mathbf{y}}_{v,C}^k]$, where $\widetilde{\mathbf{y}}_{v,c}^k$ is the probability of node $v$ belongs to class $c$ ($1\leq c \leq C$). Our goal is to use the soft label set $\{\widetilde{\mathbf{Y}}^k\in\mathbb{R}^{N\times C}\}_{k=1}^K$ from $K$ GNN models to teach an inference-friendly student model $f_w$ parameterized by $w$, such that the student model can perform better than each individual GNN. 

\subsection{Motivation: How to effectively assemble \\semi-supervised GNN models?}
\label{sec_motivation}
\begin{figure}[t]
  \centering
  \begin{subfigure}[b]{0.25\textwidth}
    \centering
    \includegraphics[width=0.99\textwidth]{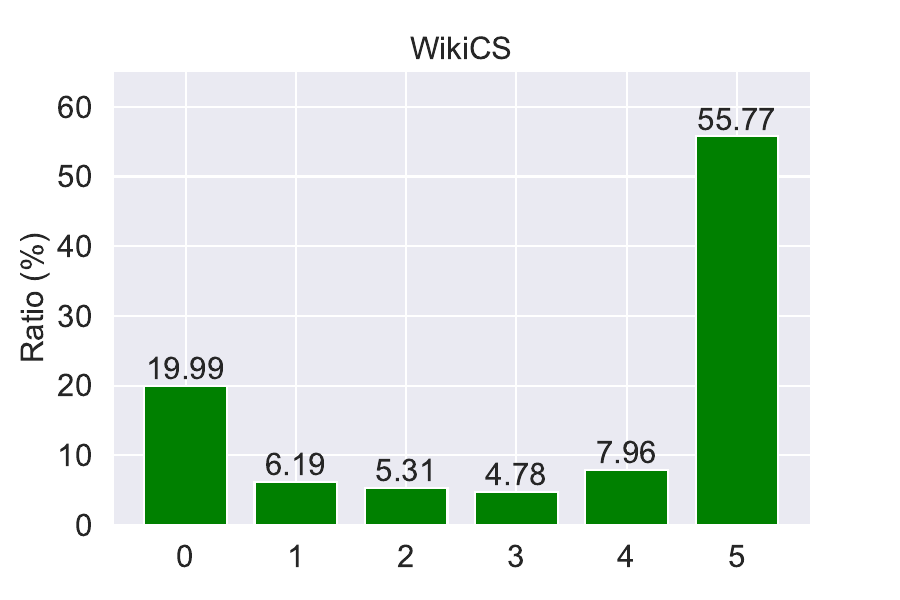}
  \end{subfigure}%
  \begin{subfigure}[b]{0.25\textwidth}
    \centering
    \includegraphics[width=0.99\textwidth]{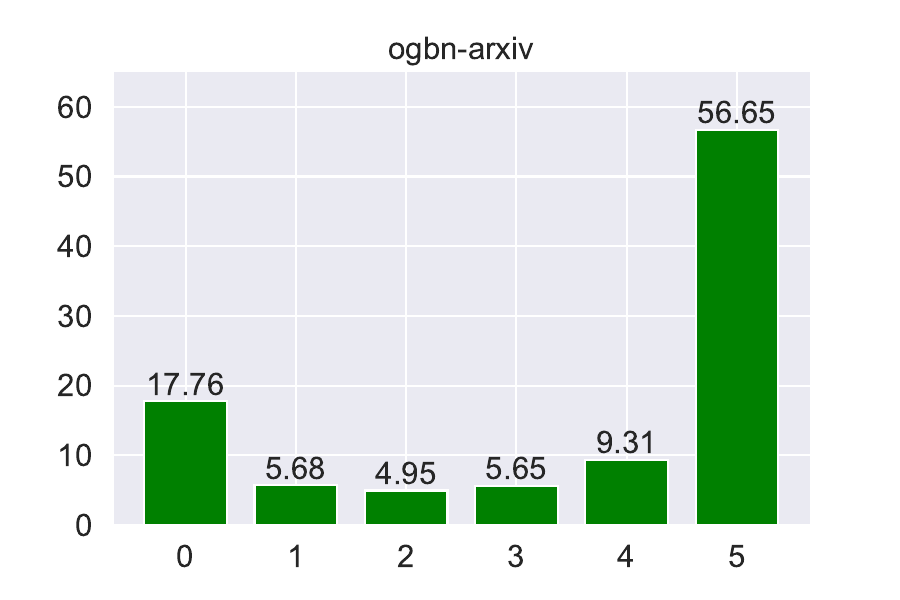}
  \end{subfigure}%
  \caption{The performance of five popular GNN models including GCN~\cite{kipf2016semi}, GAT~\cite{velivckovic2017graph}, APPNP~\cite{klicpera2018predict}, GraphSAGE~\cite{hamilton2017inductive}, and SGC~\cite{wu2019simplifying}. The x-axis indicates the number of GNN models that make the correct predictions on some nodes. The y-axis is the ratio of these nodes over all nodes in the graph. }
  \label{fig_motivation}
\end{figure}

To understand how to perform GNN ensembles effectively, we analyze the prediction statistics of multiple GNN models, following the popular semi-supervised setting~\cite{yang2021extract,hu2021ogb}. Specifically, we first pre-train five GNN models including GCN~\cite{kipf2016semi}, GAT~\cite{velivckovic2017graph}, APPNP~\cite{klicpera2018predict}, GraphSAGE (SAGE)~\cite{hamilton2017inductive}, and SGC~\cite{wu2019simplifying} using the small labeled set. Then, we apply the pre-trained GNNs to make predictions for the remaining nodes in the graph. Assuming that all nodes' ground-truth labels are available, we summarize their performance in terms of node groups in Figure~\ref{fig_motivation}. The ''2'' in the x-axis denotes the node set that is correctly classified by 2 of 5 GNN models. We make one major observation: \textbf{More than 25\% of nodes are wrongly predicted by half of GNN models.} To be specific, 31.49\% and 28.39\% nodes are wrongly predicted by more than half of GNN models (i.e., groups 0, 1, and 2) on WikiCS and ogbn-arxiv, respectively.

Given this, standard ensemble strategies like majority voting cannot boost the performance since the corresponding ensemble results are 68.51 and 71.61, which are only comparable to the best model as listed in Table~\ref{table_trans}. Furthermore, as shown in Figure~\ref{fig_toy}, GNN models tend to make wrong predictions with high certainty. For instance, while GCN is pretty confident (i.e., the color of the cell is darker) about the predictions of nodes 6 and 8, they are miss-classified (i.e., without stars). Consequently, popular ensemble heuristics, such as averaging the output, would be sub-optimal since the wrong prediction with high certainty may dominate the ensemble process, especially when the confidence for correct prediction is low.

The above observations reveal the ineffectiveness of standard ensemble learning techniques for semi-supervised GNN ensembles. It motivates us to explore a more effective ensemble learning approach for GNNs, so that the correct knowledge for each node can be effectively identified, while that incorrect knowledge from GNN models can be rejected for inference.

\subsection{The Overview of E2GNN}
Motivated by the aforementioned observations and the poor inference ability of GNNs, we suggest an alternative approach to achieve effective GNN ensembles with low latency. The key idea is to distill the knowledge of multiple GNNs into a unified student model. The overview of the proposed framework -- E2GNN is summarized in Figure~\ref{fig:e2gnn}. Given a set of pre-trained GNN models following standard semi-supervised protocol~\cite{kipf2016semi}, we first estimate the soft labels of both labeled and abundant unlabeled nodes from them. Then, we develop a node-level discriminator to control the teacher knowledge that should be mimicked by the student model. In our framework, the node-level selector is implemented by reinforcement learning, where the state is represented by the local structure and attributes of a node, and the action is the index of GNN models. Next, the student model is trained based on the agent's actions, which is a set of soft-labels of nodes that likely be correctly predicted by GNN models. Finally, the agent calculates the reward for each action to train the policy network according to the validation set, where the goal of the agent is to maximize the expected reward. This process is repeatedly iterated until convergence.

\subsection{Node-level Teacher Model Selection with Meta-Policy}
According to the findings in the motivation section~\ref{sec_motivation}, the main hurdle to performing a compelling GNN ensemble is that most GNN models wrongly predict the soft-labels of a large portion of nodes. As a result, we cannot blindly leverage the soft-labels of both labeled and unlabeled nodes to teach the student since the noisy knowledge will mislead the training. To address this, we develop a node-level teacher selector based on reinforcement learning to adaptively choose the most reliable GNN model for each node. Specifically, for each node $v$, we aim to select one of the teachers (or no teacher) for distillation. To this end, we define a meta-policy network $\pi_\theta$, parameterized by $\theta$, to make such decisions based on node-level states. The meta-policy $\pi_\theta$ takes a node-level state $s_v$ as input and outputs a probability distribution $\pi_\theta(s_v) \in \mathbb{R}^{K+1}$ for teacher sampling. We define the state and action as follows.

\textbf{State $\mathbf{s}_v$:} The state is expected to characterize the feature and structural information of the node. One possible way is to use another GNN to directly extract representations from the graph as the states. However, this will introduce additional computational costs in propagation. To minimize the costs of the meta-policy, we propose to directly use the hidden representation of the student model as the state, which enjoys two desirable properties: 1) Since the student's hidden representation already captures structural information, we can simply instantiate the meta-policy with an MLP, which avoids expensive propagation. 2) The student could learn a better hidden representation than a single GNN since it ensembles the knowledge of multiple GNN teachers.

\textbf{Action $a_v$:} The action decides which teacher model will be used for distillation for a specific node. However, it is likely that all the teachers are unreliable. For example, in Figure~\ref{fig_motivation}, all the five teacher models make wrong predictions on around 15\% to 20\% of the nodes. Thus, distilling any of the teacher models for these nodes could degrade the performance. To address this issue, we add an additional ``null'' action to allow the student not to distill any of the teacher models for a node. Specifically, we define the action space as below:
\begin{itemize}
    \item $a_v = 0$: it means we do not distill any of the teacher models. The student will simply drop the node $v$ in distillation.
    \item $a_v \in \{1, 2, ..., K\}$: it suggests we distill the knowledge of node $v$ based on the soft labels of the teacher model $k$.
\end{itemize}
In each iteration, we achieve node-level teacher selection by calculating the action probability distribution $\pi_\theta(s_v)$ for each node $v$. Based on this distribution, we then sample an action $a_v$ for each node, which will guide the distillation. The meta-policy $\pi_\theta$ will in turn be updated based on the feedback from the student model (detailed in the following subsections).

\subsection{Agent-guided Student Model Training}
Given the policy network $\pi_{\theta}$ and the soft label set $\{\widetilde{\mathbf{Y}}^k\}_{k=1}^K$ from $K$ GNN models, the student model $f_w$ is trained to mimic the correct prediction of each node determined by the agent, expressed as:
\begin{equation}
\begin{split}
    \mathcal{L} = \alpha \sum_{v\in\mathcal{V}^L} & \mathcal{L}_{CE}(f_w(\mathbf{x}_v), \mathbf{y}_v) + (1-\alpha) \sum_{v\in\mathcal{V}} \mathcal{L}_{KL}(f_w(\mathbf{x}_v), \widetilde{\mathbf{y}}^{*}_v), \\
    & \text{where} \, \, \, \ \widetilde{y}^*_v = \widetilde{\mathbf{y}}^k_v \, \,\,\,\ \text{if} \,\,\,\ k=\arg\max \pi_\theta(s_v). \\
\end{split}
\label{eq_student}
\end{equation}
$\mathcal{L}_{CE}$ denotes the cross-entropy loss on labeled nodes and $\mathcal{L}_{KL}$ is the knowledge distillation loss, such as the Kullback-Leibler (KL) divergence between the predictions of the student model and GNN teacher. Notably, $\widetilde{\mathbf{y}}^0_v$ is dynamically set to $f_w(\mathbf{x}_v)$ during the training, which means ``null'' action from the teacher since the KL divergence of the same distribution is zero. $\alpha$ is a hyper-parameter to trade off the importance of the two terms. Unlike standard knowledge distillation objective~\cite{hinton2015distilling,zhang2021graph} that utilizes all the knowledge from the teacher, Eq.~\ref{eq_student} offers the ability to choose the correct knowledge for each node among multiple teachers and can even reject the distillation. This property enables our proposal to exceed the teacher models as empirically verified in Table~\ref{table_trans} and~\ref{table_induct}.

\subsection{Optimization}
In this subsection, we introduce the optimization procedure of our framework, which iteratively updates the meta-policy $\pi_\theta$ and the student model $f_w$.

\textbf{Reward.} We use the prediction results of $f_w$ on the validation set as the reward. Let $\widetilde{\mathbf{y}}_v^s$ denote the output of $f_w$. For a sampled action $a_v$, the reward for node $v$ is defined as
\begin{equation}
    r_v =
    \begin{cases}
      -D_\text{KL}(\widetilde{\mathbf{y}}_v^{a_v} \Vert \widetilde{\mathbf{y}}_v^s), & \text{if}\ \arg\max \mathbf{y}_v = \arg\max \widetilde{\mathbf{y}}_v^{a_v} \\
      -e, & \text{otherwise}
    \end{cases}
    \label{eq_reward}
\end{equation}
where $D_\text{KL}(\cdot \Vert \cdot)$ is KL divergence, and $e$ is a constant for penalizing wrong predictions. The reward will be maximized only when the selected teacher makes the correct prediction. The KL divergence quantifies the distillation error of the node, which indicates how ''surprising'' or unexpected the teacher model is. When more than one teacher models make correct predictions, this design prioritizes the teacher model which could lead to a larger amount of learning in distillation.

\textbf{Optimization of the meta-policy.} The objective is to train the meta-policy $\pi_\theta$ such that it can maximize the expected reward:
\begin{equation}
    \mathcal{J} = \mathbb{E}[r_v],
\end{equation}
where node $v$ is any node from all the nodes in the graph. Following the policy gradient theorem~\cite{williams1992simple}, we calculate the gradient of $\mathcal{J}$ w.r.t. $\theta$ with
\begin{align}
\begin{split}
    \bigtriangledown_{\theta} \mathcal{J} & = \bigtriangledown_{\theta} \mathbb{E}[r_v] \\
    & = \mathbb{E}[r_v \bigtriangledown_\theta \log \pi_\theta(a_v|\mathbf{s}_v)],
\end{split}
\end{align}
where $a_v$ is the currently selected action for node $v$. We approximate the above gradient with the samples in the validation set:
\begin{equation}
      \bigtriangledown_\theta \mathcal{J} \approx \sum_{v \in \mathcal{V}^{\text{val}}} r_v \bigtriangledown_\theta [\log \pi_\theta(a_v|\mathbf{s}_v)],
      \label{eq:approgradient}
\end{equation}
where $\mathcal{V}^{\text{val}}$ denotes the validation set and $r_v$ is obtained based on Eq.~\ref{eq_reward} with $\mathcal{V}^{\text{val}}$. Unfortunately, the update of Eq.~\ref{eq:approgradient} can be unstable due to the high variance of the gradients. Thus, we further introduce a baseline~\cite{sutton2018reinforcement} for variance reduction, which leads to the following gradients:
\begin{equation}
      \bigtriangledown_\theta \mathcal{J} \approx \sum_{v \in \mathcal{V}^{\text{val}}} (r_v - B) \bigtriangledown_\theta [\log \pi_\theta(a_v|\mathbf{s}_v)],
       \label{eq:meta}
\end{equation}
where the baseline $B=(\sum_{v \in \mathcal{V}^{\text{val}}} r_v) / |\mathcal{V}^{\text{val}}|$ is the mean reward across the nodes in the meta-set, and $(r_v - B)$ can be interpreted as the advantage, i.e., the extent to which the current selected teacher model is better than the average reward that can be achieved by all the teacher models. In practice, we can sample actions in batches to approximate the above gradient in training.

\textbf{Overall optimization.} Following~\cite{liu2018darts,you2021graph}, we iteratively update the meta-policy $\pi_\theta$ and the student model $f_w$ towards convergence, which is summarized in Algorithm~\ref{alg:example}.

\begin{algorithm}[tb]
\vspace{-1pt}
   \caption{Optimization of E2GNN}
   \label{alg:example}
   \KwIn{Initial meta-policy $\pi_\theta$ and student model $f_w$}
   \While{not converge }{
   1. Obtain the hidden representation of $f_w$ for the unlabeled nodes. \\
   2. Train $\pi_\theta$ with policy gradient based on Eq.~\ref{eq:meta} using the validation set.\\
   4. Fix $\pi_\theta$ and update $f_w$ based on Eq.~\ref{eq_student}.}
\textbf{Return} The student model $f_w$.
\vspace{-3pt}
\end{algorithm}

\subsection{More Discussions on E2GNN}
\textbf{E2GNN is applicable to other student models.} While we illustrate the idea of E2GNN using MLPs as the student backbone for efficiency purposes, it can be easily applied to other graph models, such as GNNs~\cite{wu2020comprehensive}, if the latency requirement is satisfied. We analyze the applicability of E2GNN on GNN students in Table~\ref{table_gnn}.

\noindent\textbf{Relation to multi-teacher knowledge distillation works.} 
Recently, some knowledge distillation efforts have been proposed in computer vision~\cite{asif2019ensemble,zhu2018knowledge} and natural language processing~\cite{yuan2021reinforced} domains to compress a set of deep neural networks, such as CNN~\cite{krizhevsky2017imagenet} and BERT~\cite{devlin2018bert}, into a lightweight student model. While these methods are conceptually similar to ours, they are designed for Euclidean data, such as images and text, while we work on graph-structured data. More importantly, they do not consider the unreliable issue of teachers, i.e., the wrongly predicted samples, and aim to select at least one teacher model for each sample. Conversely, we aim to identify the most predictive teacher model for each node and sometimes even allow to reject distillation from teacher GNNs.

\section{Experiments}
In this section, we conduct extensive experiments to benchmark the effectiveness and efficiency of E2GNN. Throughout the experiments, we try to answer the following research questions:
\begin{itemize}
    \item \textbf{RQ1:} What does the proposed meta-policy actually do? Why is it beneficial for the student model training?
    \item \textbf{RQ2:} How effective is the proposed E2GNN compared with state-of-the-art node classification baselines in both transductive and inductive scenarios? 
    \item \textbf{RQ3:} How does each component of E2GNN contribute to the performance?
    \item \textbf{RQ4:} What are the impacts of feature and topology perturbations on E2GNN? 
    \item \textbf{RQ5:} How efficient is E2GNN compared with other GNN ensemble baselines?
\end{itemize}

\begin{table}[ht]
\centering
  \caption{Dataset statistics.}
  \begin{small}
\setlength{\tabcolsep}{2.5pt}
  {
  \begin{tabular}{c| c |c |c | c}
   \toprule
     Data&\# Nodes &\# Edges &\# Features &\# Classes\\
     \hline
     Cora &$2,708$ &$5,429$ &$1,433$ &$7$\\
     CiteSeer &$3,312$ &$4,660$ &$3,703$ &$6$ \\
     PubMed &$19,717$ &$44,338$ &$500$ &$3$\\
     Wiki-CS &$11,701$ &$ 216,123$ &$300$ &$10$ \\
     Amazon-Computers &$13,752$ &$245,861$ &$767$ &$10$\\
     Amazon-Photo &$7,650$ &$119,081$ &$745$ &$8$\\
     Coauthor-CS &$18,333$ &$81,894$ &$6,805$ &$15$\\
     ogbn-arxiv &$169,343$ &$1,166,243$ &128&$40$\\
 \bottomrule
\end{tabular} }
\end{small}
\label{dataset_stat}
\end{table}

\begin{table}[ht]
\centering
\caption{Hyperparameters of different GNN teachers on small datasets: Cora, CiteSeer, PubMed, WikiCS, Amazon-Computers, Amazon-Photo, and Coauthor-CS.}
\label{table_hyper_small}
\begin{small}
\begin{tabular}{lcccccc}
\toprule
 &SAGE &GCN &GAT &APPNP &SGC\\
\midrule
\# layers & 2 & 2 & 2 & 2 &2 \\
hidden dim &128 & 128 & 128 &128 &128 \\  
learning rate &0.01 & 0.01 & 0.01 & 0.01 &0.01 \\
weight decay &0.001 &0.0005 &0.0005 &0.0005 &0.001 \\
dropout &0.5 &0.5 &0.5 &0.0 &0.0 \\
attention heads &- & - &8 &- &- \\
power iterations & - &- &- & 10 &- \\
\bottomrule
\end{tabular}
\end{small}
\end{table}

\begin{table}[ht]
\centering
\caption{Hyperparameters of different GNN teachers on ogbn-arxiv.}
\label{table_hyper_arxiv}
\begin{small}
\begin{tabular}{lcccccc}
\toprule
 &SAGE &GCN &GAT &APPNP &SGC\\
\midrule
\# layers & 3 & 3 & 3 & 3 &3 \\
hidden dim &256 & 256 & 256 &256 &256 \\  
learning rate &0.01 & 0.01 & 0.01 & 0.01 &0.01 \\
weight decay &0.0 &0.0 &0.0 &0.0 &0.0 \\
dropout &0.5 &0.5 &0.5 &0.5 &0.5 \\
attention heads &- & - &8 &- &- \\
power iterations & - &- &- & 10 &- \\
\bottomrule
\end{tabular}
\end{small}
\end{table}

\begin{table}[ht]
\centering
\caption{Hyperparameters of E2GNN for MLP student.}
\label{table_hyper_mlp}
\resizebox{\linewidth}{!}{
\begin{tabular}{lcccccc}
\toprule
&\# layers &hidden dim &learning rate &weight decay &dropout\\
\midrule
Cora &2 &128 &0.008 &0.005 &0.5\\
CiteSeer &2 &128 &0.001 &0.01 &0.6\\  
PubMed &2&128&0.001&0.005&0.3\\
WikiCS &2&128&0.008&0&0.5\\
Computer &2&128&0.005&0.001&0.5\\
Photo &2&128&0.01&0&0.5\\
CS &2&128&0.008&0.01&0.5\\
Arxiv &3&1024&0.001&0.001&0.5\\
\bottomrule
\end{tabular}
}
\end{table}

\subsection{Experimental Setup}
\subsubsection{Datasets.}
In our experiments, we use eight benchmark datasets including three Planeteoid datasets~\cite{sen2008collective} (\textbf{Cora}, \textbf{CiteSeer}, and \textbf{PubMed}), two co-purchase networks from Amazon~\cite{mcauley2015image} (Amazon-Computer (\textbf{Computer}) and Amazon-Photo (\textbf{Photo})), one co-authorship graph from Microsoft Academic Graph from the KDD Cup 2016 challenge~\cite{sinha2015overview} (Coauthor-CS (\textbf{CS})), and one reference networks constructed based on Wikipedia~\cite{mernyei2020wiki} (\textbf{WikiCS}), and a large OGB dataset~\cite{hu2020open} (ogbn-Arxiv (\textbf{Arxiv})). We summarize the data statistics in Table~\ref{dataset_stat}.

To provide a fair comparison between different models, we follow the standard dataset splits and training procedure. Specifically, for Planetoid datasets, we follow the common semi-supervised practice~\cite{kipf2016semi} to generate the training/validation/testing sets. For Computer, Photo, WikiCS, and CS, we follow~\cite{yang2021extract,zhang2021graph} to generate the semi-supervised data splits. For ogbn-arxiv, we adopt the default splits in~\cite{hu2021ogb}. 

\begin{figure*}[htbp]
  \centering
  \begin{subfigure}[b]{0.33\textwidth}
    \centering
    \includegraphics[width=0.9\textwidth]{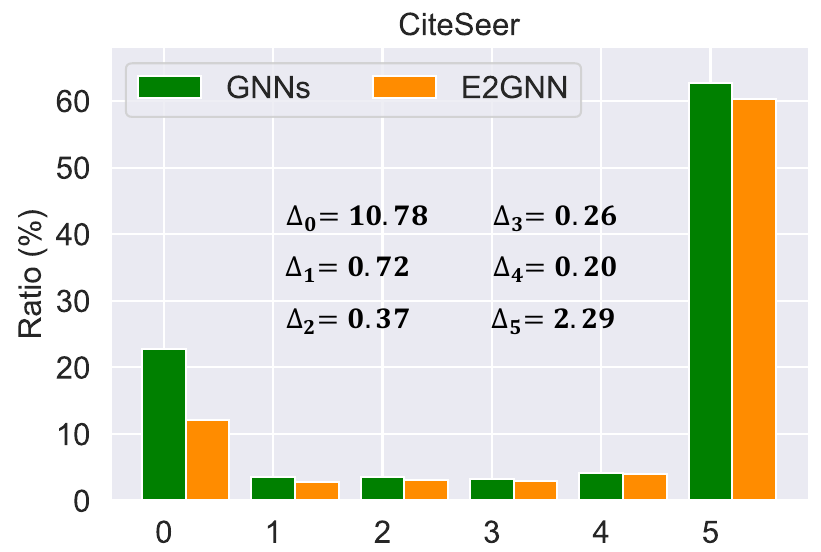}
  \end{subfigure}%
  \begin{subfigure}[b]{0.33\textwidth}
    \centering
    \includegraphics[width=0.9\textwidth]{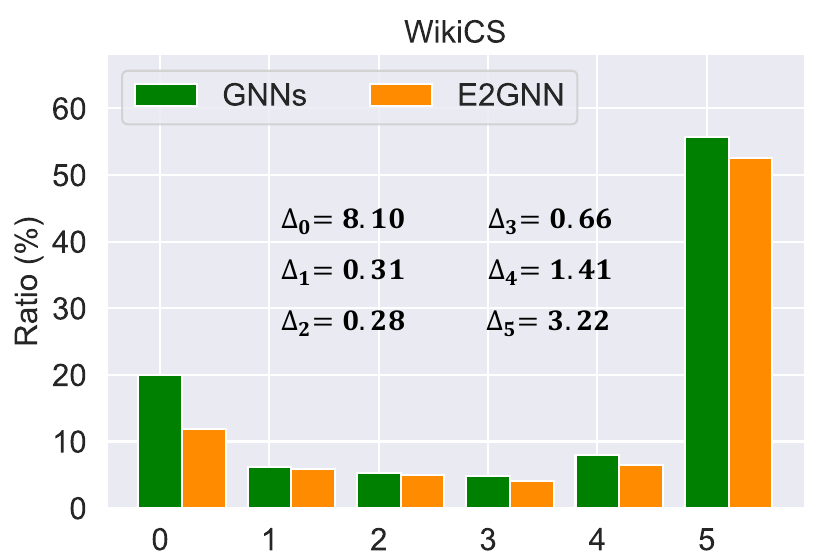}
  \end{subfigure}%
  \begin{subfigure}[b]{0.33\textwidth}
    \centering
    \includegraphics[width=0.9\textwidth]{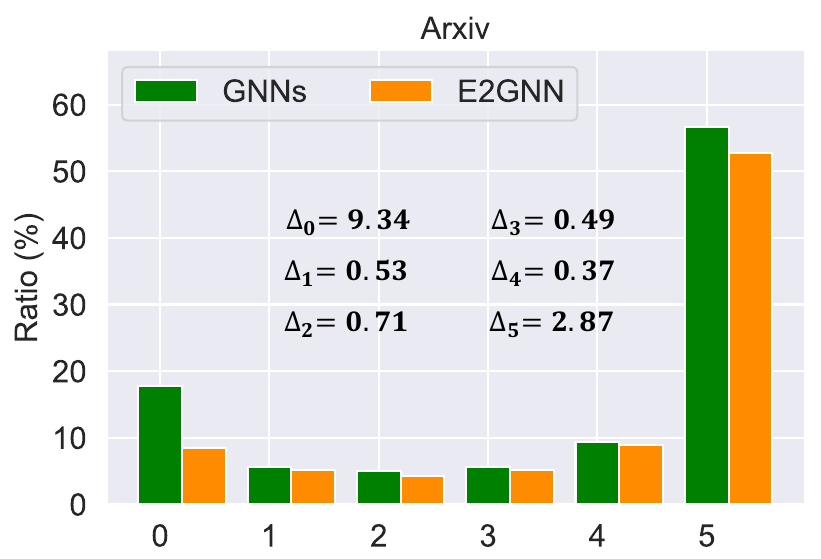}
  \end{subfigure}%
  \caption{Visualization of the meta-policy's decisions. The x-axis indicates six groups of nodes, where the value means the number of teacher GNNs that can correctly predict the nodes in the corresponding group. The green bar represents the ratio of the nodes in a group among all the nodes. The orange bar indicates the ratio that the meta-policy makes a good decision (i.e., the selected teacher makes correct predictions for groups 1-5 or take the ''null'' action for group 0). $\Delta_{*}$ indicates the performance gap between GNN models and E2GNN under different groups, which is the lower the bette. For example, $\Delta_2$ means the ratio of nodes which are correctly predicted by 2 GNN models yet filtered out by E2GNN. Obviously, E2GNN can sample nodes that are correctly predicted by at least one GNN model while rejecting nodes being wrongly classified by all GNNs, i.e., group ''0''.    
  }
  \label{fig_case}
\end{figure*}

\subsubsection{The architecture of E2GNN}
\noindent\textbf{Teacher GNNs.} We consider the following five popular GNN models as teacher backbone. For these methods, we follow~\cite{yang2021extract} to set their hyperparameters. 
\begin{itemize}
    \item \textbf{GCN}~\cite{kipf2016semi} is the pioneering work that introduces convolution operators on graph-structured data to tackle the semi-supervised classification problem. 
    \item \textbf{GAT}~\cite{velivckovic2017graph} improves GCN by aggregating neighboring nodes with different weights learned by attention mechanism. 
    \item \textbf{APPNP}~\cite{klicpera2018predict} improves GCN by balancing the preservation of local information and the use of a wide range of neighbor information. 
    \item GraphSAGE~\cite{hamilton2017inductive} (\textbf{SAGE}) upgrades GCN to the inductive scenario by aggregating a set of sampled local neighbors. 
    \item \textbf{SGC}~\cite{wu2019simplifying} reduces the extra complexity of GCN by removing the non-linearity between GCN layers and compressing the model weights.  
\end{itemize}

\noindent\textbf{Student model.} The default student in E2GNN is simple MLPs. The architecture of MLPs is the same with the teacher GNN, i.e., the number of layers are the same. 

\subsubsection{Competitors}
We consider two categories of baselines. One is the conventional GNN models, including GCN~\cite{kipf2016semi}, GAT~\cite{velivckovic2017graph}, APPNP~\cite{klicpera2018predict}, GraphSAGE~\cite{hamilton2017inductive}, and SGC~\cite{wu2019simplifying}. These GNN methods have shown promising results over a variety of graph applications~\cite{zhou2020graph,wu2020graph}. The other one is state-of-the-art ensemble methods. Their introduction is shown below:


\begin{itemize}
    \item \textbf{GNNE}~\cite{kosasih2021graph} is a classic ensemble approach on GNNs by averaging the outputs of different GNN models equally. It does not consider the inference issue of GNN ensembles.
    \item \textbf{GEENI}~\cite{nagarajan2022efficient} improves the efficiency of GNNE by reducing the edge connections and transformation computation via edge pruning and weight pruning, respectively.
    \item \textbf{EKD-U} is our implementation of the equal weight method~\cite{allen2020towards}. It computes the distillation knowledge by averaging the outputs of GNN models equally, and then the student model learns to mimic the aggregated teacher knowledge. 
    \item \textbf{EKD-W} is our implementation of the adaptive weighting method~\cite{yuan2021reinforced}. It selectively combines the outputs of all GNN models, and then the student model learns to mimic the integrated knowledge. 
    \item \textbf{GLNN}~\cite{zhang2021graph} is originally proposed for single teacher model. We extend it to multiple teachers by using the output of the best performed teacher for distillation.
\end{itemize}

\subsubsection{Implementation details}
\noindent\textbf{Evaluation Setting.} We evaluate the performance of our model in both transductive and inductive scenarios. Given a graph $\mathcal{G}$, with feature matrix $\mathbf{X}$ and label matrix $\mathbf{Y}$, the standard transductive setting also uses the testing nodes to construct the adjacency matrix during training. For inductive setting, following the practice in~\cite{zhang2021graph}, we fist randomly sampling some nodes to construct the testing only set $\mathcal{V}_{ind}^U\in\mathcal{V}^U$, resulting in two disjoint unlabeled node set: inductive node set $\mathcal{V}_{ind}^U$ and observed node set $\mathcal{V}_{obs}^U$. The difference between the $\mathcal{V}_{ind}^U$ and $\mathcal{V}_{obs}^U$ is that nodes in $\mathcal{V}_{ind}^U$ will not be used during the training, while nodes in $\mathcal{V}_{obs}^U$ will be implicitly used to construct the adjacency matrix. In a summary, the two evaluation settings are tested as follows: 
\begin{itemize}
    \item \textbf{Transductive:} the model is trained on $\mathcal{G}$, $\mathbf{X}$, and $\mathbf{Y}^L$, and tested on $\mathbf{X}^U$, and $\mathbf{Y}^U$, with KD based on $\mathbf{z}_v$ for $v\in\mathcal{V}$. 
    \item \textbf{Inductive:} the model is trained on $\mathcal{G}_{obs}$, $\mathbf{X}_{obs}^U$, $\mathbf{X}^L$, and $\mathbf{Y}^L$, and tested on $\mathbf{X}_{ind}^U$ and $\mathbf{Y}_{ind}^L$, with KD based on $\mathbf{z}_v$ for $v\in\mathcal{V}_{obs}^U\cup\mathcal{V}^L$. 
\end{itemize}

\noindent\textbf{Hyperparameter.} We conduct experiments based on the benchmark PyG~\cite{fey2019fast} library, where all GNN models considered in this paper are provided. For all GNN models, following common practice in~\cite{zhang2021graph,yang2021extract,hu2021ogb}, we employ three GNN layers with dimension 256 for ogbn-arxiv dataset, and two GNN layers with dimension 128 for other datasets. For KD methods including EKD-U, EKD-W, GLNN, and E2GNN, we adopt three-layer MLPs with dimension 1024 as the student model for ogbn-arxiv, while two-layer MLPs with dimension 128 for other datasets. We found that E2GNN is not sensitive to the trade-off parameter $\alpha$ as shown in Figure~\ref{fig_parameter}, so we set $\alpha=0.0$ if not specified. The penalty for wrong prediction $e$ is set to 5. To avoid randomness, all the experiments are independently run 10 times and the mean and the standard deviation results are reported. Tables~\ref{table_hyper_small} and~\ref{table_hyper_arxiv} report the detailed configurations. For the student model, the detailed hyperparameter settings for different datasets are summarized in Table~\ref{table_hyper_mlp}.

\noindent\textbf{Hardware.}
We conduct all the experiments on a server with 48 Intel(R) Xeon(R) Silver 4116 CPU @ 2.10GHz processors, 188 GB memory, and four NVIDIA GeForce RTX 3090 GPUs.

\subsection{What is E2GNN doing? A Case Study (\textbf{RQ1})}
To investigate what E2GNN actually does on real-world datasets, we check the agent's action and visualize the results according to six groups in Figure~\ref{fig_case}. We make three major observations. \textbf{First}, E2GNN can effectively spot nodes being correctly classified by the majority of GNN teachers, i.e., nodes in groups 3, 4, and 5. Taking Arxiv as an example, the performance gap between the optimal GNN ensembles and E2GNN are 0.49\%, 0.37\%, and 2.87\% in groups 3, 4, and 5, respectively. That is, only 3.73\% nodes are wrongly filtered out by our policy, which is relatively small given that the overall nodes in the three groups are 71.61\% as shown in Figure~\ref{fig_motivation}. \textbf{Second}, E2GNN benefits in picking up nodes that are correctly classified by the minority of GNN models, i.e., nodes in groups 1 and 2. Specifically, the performance gap between optimal GNN ensembles and E2GNN on Arxiv is 0.53\% and 0.71\% in groups 1 and 2, respectively. According to Figure~\ref{fig_motivation}, it means E2GNN can correctly identify 9.39\% nodes belonging to the two challenging groups. It is noteworthy that nodes in the two groups are difficult for vanilla ensemble strategies, e.g., average rule or majority voting, to make correction predictions. We believe this merit is one of the key factors that lead to the success of our model. \textbf{Third}, E2GNN can reduce the impact of noisy nodes for distillation. In Figure~\ref{fig_case}, E2GNN reduces 12.04\%, 11.88\%, and 8.42\% nodes belong to group 0 on CiteSeer, WikiCS, and PubMed datasets, respectively. This result is promising since nodes in group 0 are the wrong guidance for the student model. We believe the ability to filter out these noisy nodes is another key factor for the success of E2GNN. In summary, E2GNN works by filtering out the noisy nodes in group 0 while preserving nodes in other groups for distillation. These observations shed light on the effectiveness of E2GNN in assembling GNNs.

\begin{table*}[ht]
\centering
  \caption{Node classification accuracy on transductive scenarios. GEENI results is different from those reported in the original paper, because we use the common GNNs teachers without specific configuration. }
\setlength{\tabcolsep}{2.0pt}
  {
    \begin{tabular}{l c| c c c c c c c c c}
    \toprule
     Teacher &Student &Cora &CiteSeer &PubMed &WikiCS &Compute &Photo &CS &Arxiv\\
     \midrule
    \multirow{5}*{Teacher}
    &GCN &$82.88\pm0.28$ &$69.30\pm0.28$ &$78.30\pm0.27$ &$67.71\pm0.64$ &$80.78\pm0.67$ &$91.43\pm0.49$ &$90.09\pm0.25$ &$72.06\pm0.39$ \\
    &SAGE &$79.56\pm0.52$ &$69.48\pm0.27$ &$76.81\pm0.29$ &$65.79\pm0.88$ &$83.74\pm2.16$ &$90.74\pm0.84$ &$90.50\pm0.38$ &$71.61\pm0.16$\\
    &GAT &$81.23\pm0.63$ &$69.81\pm0.42$ &$77.92\pm0.25$ &$70.03\pm0.97$ &$82.12\pm1.51$ &$92.32\pm0.74$ &$89.71\pm0.72$ &$70.98\pm0.37$\\
    &APPNP &$81.96\pm0.64$ &$69.74\pm0.21$ &$79.05\pm0.19$ &$68.68\pm1.08$ &$83.61\pm1.21 $ &$90.16\pm0.95$ &$91.65\pm0.28$ &$66.31\pm1.03$\\
    &SGC &$81.91\pm0.47$ &$69.15\pm0.31$ &$78.21\pm0.14$ &$68.36\pm0.59$ &$81.36\pm0.64$ &$91.61\pm0.51$ &$90.13\pm0.38$ &$70.05\pm0.42$\\
   \midrule
    \multirow{6}{*}{Student}
    &GLNN &$82.55\pm0.84$ &$69.98\pm0.32$ &$80.27\pm0.36$ &$70.69\pm1.17$ &$83.05\pm0.72$ &$92.12\pm0.67$ &$91.19\pm0.52$ &$71.89\pm0.37$\\
    &GNNE &$82.11\pm0.34$ &$70.25\pm0.28$ &$79.53\pm0.29$ &$70.23\pm0.22$ &$83.27\pm0.29$ &$92.37\pm0.37$ &$91.57\pm0.28$ &$68.75\pm1.10$\\
    &GEENI &$82.36\pm0.28$ &$70.30\pm0.33$ &$79.43\pm0.24$ &$68.79\pm0.57$ &$83.37\pm0.24$ &$92.11\pm0.31$ &$90.89\pm0.35$ &$70.08\pm0.23$\\
    &EKD-U &$81.98\pm0.82$ &$69.59\pm0.30$ &$79.45\pm0.35$ &$69.08\pm0.48$ &$82.77\pm0.55$ &$92.69\pm0.22$ &$91.64\pm0.36$ &$70.36\pm0.31$\\
    &EKD-W &$82.28\pm0.50$ &$70.57\pm0.42$ &$80.34\pm0.29$ &$70.92\pm0.53$ &$83.16\pm0.23$ &$92.39\pm0.49$ &$91.44\pm0.44$ &$71.67\pm0.36$\\
    \cmidrule(r){2-10}
    &E2GNN &$\bf83.53\pm0.19$ &$\bf72.10\pm0.20$ &$\bf81.85\pm0.45$ &$\bf73.23\pm0.43$ &$\bf84.81\pm0.12$ &$\bf93.10\pm0.78$ &$\bf93.06\pm0.22$ &$\bf72.51\pm0.13$\\
  \bottomrule
\end{tabular}}
\label{table_trans}
\end{table*}

\begin{table*}[t]
\centering
  \caption{Node classification accuracy on inductive scenarios. GEENI results is different from those reported in the original paper, because we use the common GNNs teachers without specific configuration. }
\setlength{\tabcolsep}{2.0pt}
  {
    \begin{tabular}{l c| c c c c c c c c c}
    \toprule
     Teacher &Student &Cora &CiteSeer &PubMed &WikiCS &Compute &Photo &CS &Arxiv\\
     \midrule
    \multirow{5}*{Teacher}
    &GCN &$80.84\pm0.19$ &$71.90\pm0.45$ &$79.40\pm0.31$ &$65.53\pm0.25$ &$80.33\pm0.45$ &$90.12\pm0.37$ &$90.19\pm0.38$ &$71.72\pm0.24$\\
    &SAGE &$80.60\pm0.33$ &$71.16\pm0.24$ &$76.40\pm0.75$ &$66.29\pm1.33$ &$81.84\pm1.15$ &$90.47\pm0.28$ &$90.81\pm0.22$ &$71.62\pm0.09$ \\
    &GAT &$81.78\pm0.69$ &$72.46\pm0.37$ &$78.00\pm0.26$ &$69.06\pm1.08$ &$80.39\pm1.22$ &$91.60\pm0.97$ &$89.41\pm0.65$ &$70.53\pm0.22$\\
    &APPNP &$82.09\pm0.28$ &$72.27\pm0.67$ &$79.30\pm0.12$ &$67.18\pm0.78$ &$81.93\pm0.48$ &$89.70\pm1.05$ &$91.84\pm0.33$ &$65.95\pm0.87$\\
    &SGC &$80.84\pm0.30$ &$71.34\pm0.44$ &$78.80\pm0.59$ &$65.27\pm0.36$ &$79.76\pm1.35$ &$88.29\pm0.65$ &$88.88\pm0.78$ &$70.04\pm0.33$ \\
   \midrule
    \multirow{6}{*}{Student}
    &GLNN &$81.68\pm0.68$ &$72.75\pm1.08$ &$79.88\pm0.23$ &$70.10\pm1.41$ &$81.59\pm0.26$ &$91.31\pm1.08$ &$91.38\pm0.53$ &$70.67\pm0.54$\\
    &GNNE &$81.79\pm0.38$ &$72.68\pm0.82$ &$79.17\pm0.31$ &$69.42\pm0.55$ &$81.64\pm0.47$ &$91.54\pm0.28$ &$91.09\pm0.49$ &$68.25\pm0.77$\\
    &GEENI &$81.86\pm0.17$ &$72.56\pm0.55$ &$79.32\pm0.43$ &$68.23\pm0.45$ &$81.70\pm0.38$ &$91.32\pm0.37$ &$90.41\pm0.51$ &$69.88\pm0.49$\\
    &EKD-U &$80.77\pm0.78$ &$72.36\pm0.45$ &$79.28\pm0.34$ &$69.74\pm0.58$ &$81.53\pm0.30$ &$91.57\pm0.41$ &$91.45\pm0.29$ &$70.35\pm0.26$\\
    &EKD-W &$81.92\pm0.30$ &$72.69\pm0.55$ &$80.12\pm0.33$ &$70.31\pm0.34$ &$81.66\pm0.53$ &$91.38\pm0.44$ &$91.28\pm0.34$ &$71.21\pm0.43$\\
    \cmidrule(r){2-10}
    &E2GNN &$\bf82.79\pm0.52$ &$\bf74.24\pm0.71$ &$\bf81.07\pm0.27$ &$\bf71.02\pm0.56$ &$\bf83.29\pm0.49$ &$\bf92.56\pm0.37$ &$\bf92.35\pm0.30$ &$\bf72.21\pm0.26$\\
  \bottomrule
\end{tabular}}
\label{table_induct}
\end{table*}

\subsection{How Effective is E2GNN Compared with SOTA Methods? (RQ2)}
To evaluate the effectiveness of E2GNN against state-of-the-art baselines, we conduct experiments on both transductive and inductive scenarios and report the classification results on Tables~\ref{table_trans} and~\ref{table_induct} accordingly. We have three observations. \textbf{First}, E2GNN achieves improved classification performance over five GNN teachers across two evaluation scenarios in general. Specifically, in Table~\ref{table_trans}, E2GNN improves the best GNN teacher 0.78\%, 3.28\%, 2.27\%, 4.56\%, 1.28\%, 0.84\%, 1.54\%, and 0.62\% on Cora, CiteSeer, PubMed, WikiCS, Computer, Photo, CS, and Arxiv, respectively. \textbf{Second}, compared with single teacher distillation method-GLNN, E2GNN consistently performs better in 16 evaluation cases. Although GLNN is trained based on the best GNN teacher, it fails to capture the complementary capacity among different GNNs. These results demonstrate our motivation to explore ensemble learning for GNN models. \textbf{Third}, compared with four ensemble baselines (GNNE, GEENI, EKD-U and EKD-W), E2GNN performs better on both transductive and inductive settings. The main difference between them is that our model can reject distillation from nodes that are wrongly predicted by all GNN models while preserving challenging nodes that are only predicted by a minority of GNN models, i.e., groups 0, 1, and 2 in Figure~\ref{fig_case}. This comparison validates the effectiveness of our proposal in integrating GNN models to achieve improved results.

\begin{table}[t]
\centering
  \caption{Ablation studies of E2GNN on the transductive setting.}
  \begin{small}
\setlength{\tabcolsep}{3.5pt}
  {
    \begin{tabular}{l c c c c }
    \toprule
        &E2GNN &E2GNN-null &E2GNN-rand \\
     \midrule
     WikiCS &$73.23\pm0.43$
     &$70.81\pm0.42$
     &$69.98\pm0.78$\\
     Computer &$84.81\pm0.12$
     &$83.34\pm0.57$
     &$82.78\pm0.56$\\
     CS &$93.06\pm0.22$
     &$91.58\pm0.27$
     &$90.30\pm0.44$\\
     Arxiv &$72.51\pm0.13$
     &$71.97\pm0.23$
     &$70.46\pm0.25$\\
  \bottomrule
\end{tabular}}
\end{small}
\label{ablation_study}
\end{table}


\subsection{Ablation Study (RQ3)} 
To evaluate the contributions of several components in E2GNN, we introduce two variants: E2GNN-null and E2GNN-rand. E2GNN-null is obtained by excluding the null action. E2GNN-rand is obtained by replacing the meta-policy with a random search. Due to limited space, we select 4 relatively large datasets for this study and report the results in Table~\ref{ablation_study}. From the table, we can see that E2GNN consistently performs better than E2GNN-null, which indicates the importance of null action in our meta-policy. Moreover, E2GNN outperforms E2GNN-rand over four cases with a wide margin, which demonstrates our motivation to design a learnable filter via reinforcement learning. The above observations verify the necessity and effectiveness of the proposed meta-policy design. 

\begin{table*}[ht]
\centering
  \caption{The impact of student models on E2GNN.}
\setlength{\tabcolsep}{2.0pt}
  {
    \begin{tabular}{l c| c c c c c c c c c}
    \toprule
      &Student &Cora &CiteSeer &PubMed &WikiCS &Compute &Photo &CS &Arxiv\\
     \midrule
    \multirow{2}{*}{E2GNN}
    &MLPs &$83.53\pm0.19$ &$72.10\pm0.20$ &$81.85\pm0.45$ &$73.23\pm0.43$ &$84.81\pm0.12$ &$93.10\pm0.78$ &$93.06\pm0.22$ &$72.51\pm0.13$\\
    &GCN &$83.63\pm0.06$ &$72.09\pm0.15$ &$81.71\pm0.13$ &$72.30\pm0.28$ &$85.72\pm0.18$ &$92.88\pm0.56$ &$92.13\pm0.02$ &$73.01\pm0.04$\\
  \bottomrule
\end{tabular}}
\label{table_gnn}
\end{table*}

\begin{table}[ht]
\centering
  \caption{Inference time (in milliseconds) of E2GNN and baseline methods on typical datasets.}
\setlength{\tabcolsep}{2.0pt}
  {
    \begin{tabular}{l c c c c c| c c |c c c}
    \toprule
      &GCN &SAGE &GAT &APPNP &SGC &GNNE &GEENI &E2GNN\\
     \midrule
    CS &$3.51$ &$3.17$ &$4.36$ &$5.05$ &$2.61$ &$19.94$ &$10.33$ &$\bf1.71$\\
    Arxiv &$28.79$ &$20.57$ &$70.36$ &$10.40$ &$5.20$ &$137.55$ &$100.24$ &$\bf1.30$\\
  \bottomrule
\end{tabular}}
\label{table_time}
\end{table}

\subsection{Can E2GNN Improve the Robustness of GNNs? (RQ4)}
\begin{figure}[t]
  \centering
  \begin{subfigure}[b]{0.25\textwidth}
    \centering
    \includegraphics[width=0.98\textwidth]{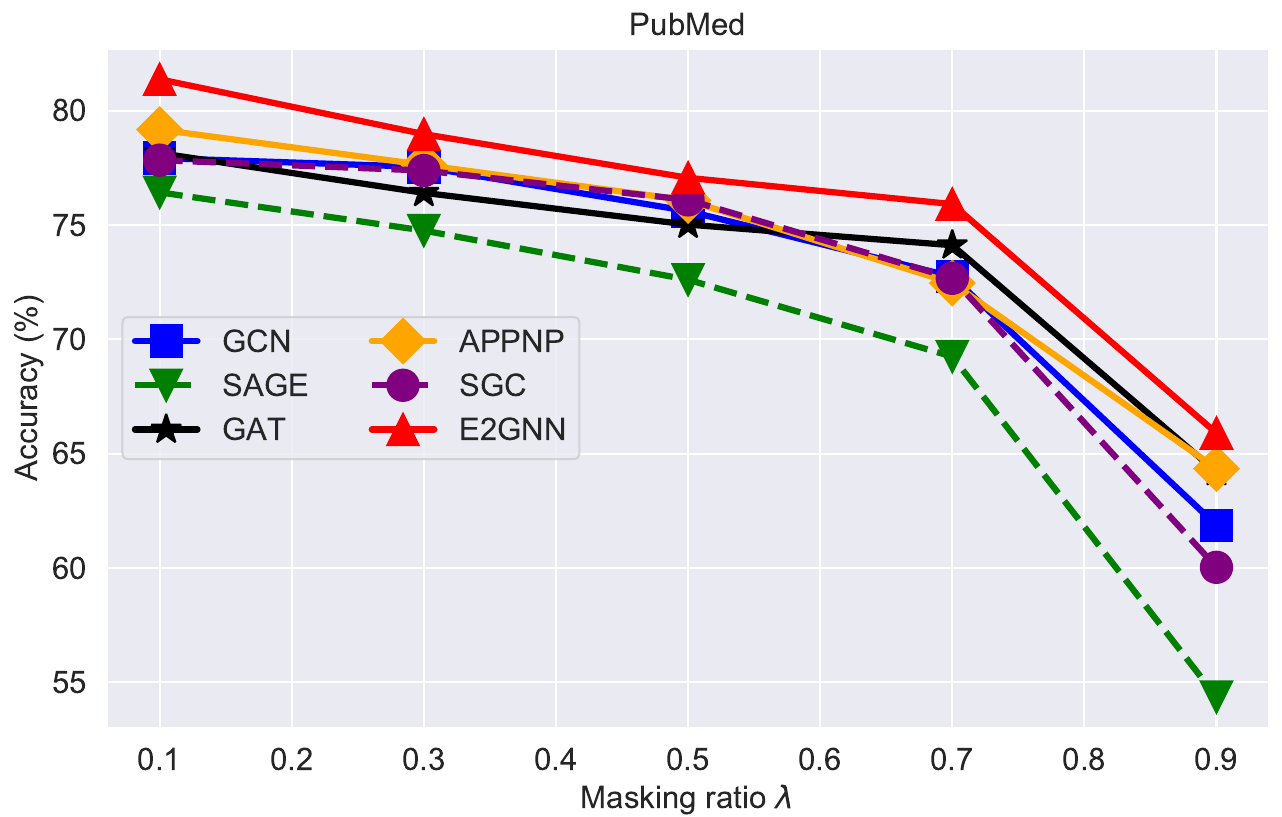}
  \end{subfigure}%
  \begin{subfigure}[b]{0.25\textwidth}
    \centering
    \includegraphics[width=0.98\textwidth]{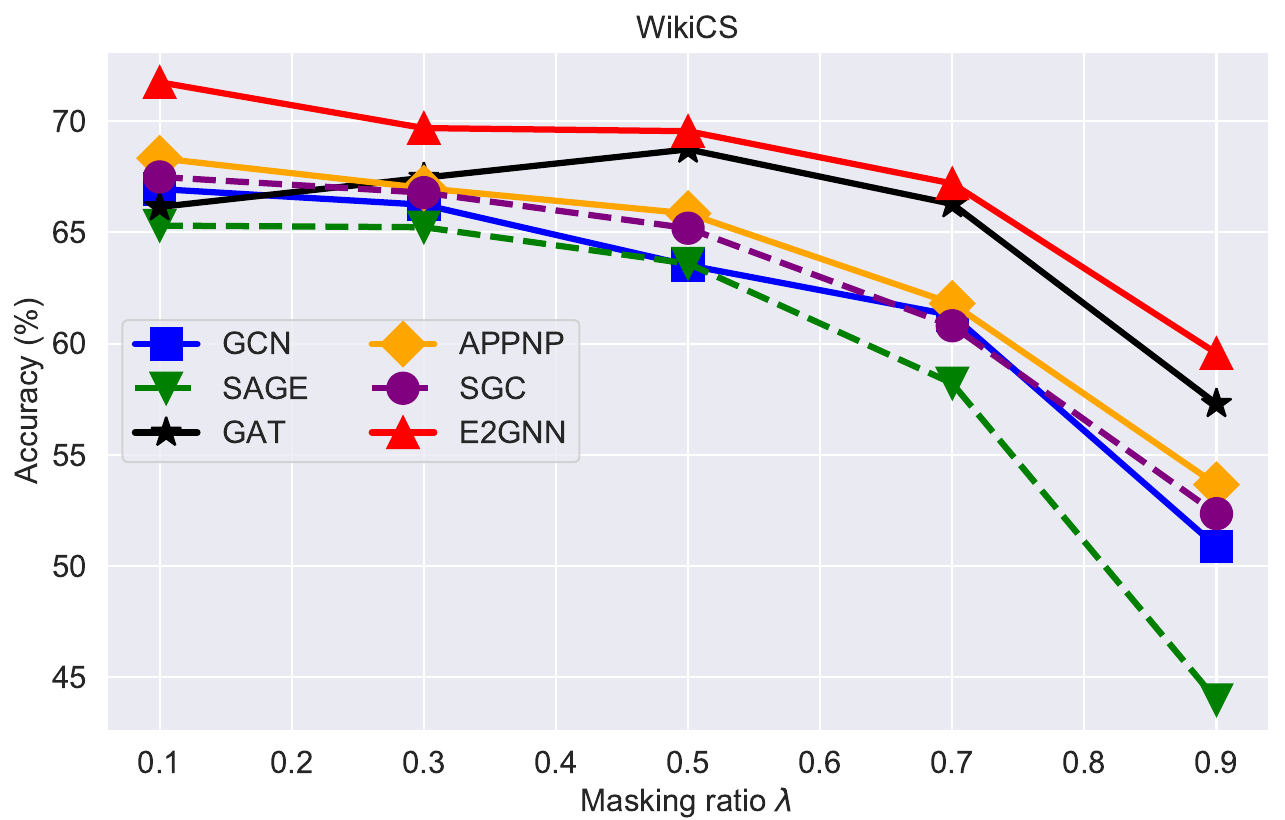}
  \end{subfigure}%
  \caption{Robust analysis of E2GNN on feature masking. }
  \label{fig_robust_fm}
\end{figure}

\begin{figure}[t]
  \centering
  \begin{subfigure}[b]{0.25\textwidth}
    \centering
    \includegraphics[width=0.98\textwidth]{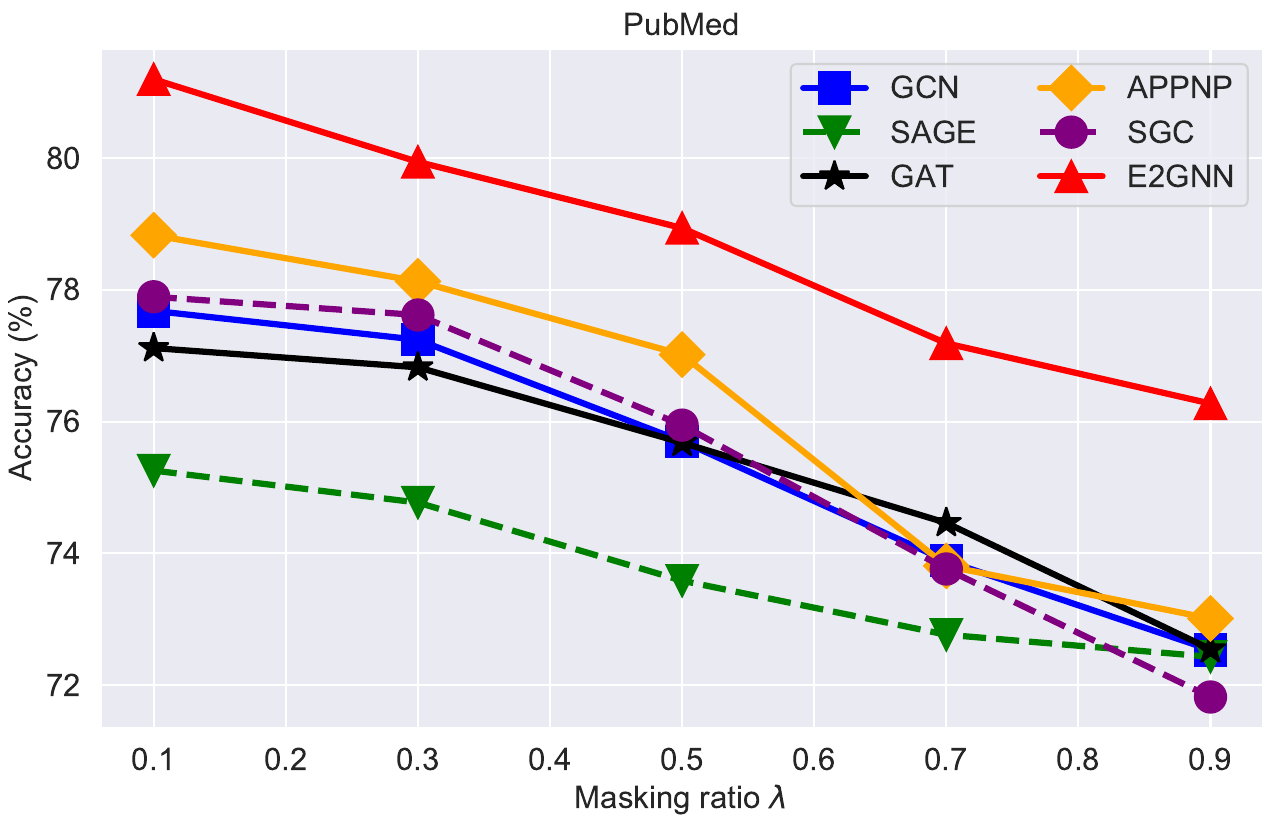}
  \end{subfigure}%
  \begin{subfigure}[b]{0.25\textwidth}
    \centering
    \includegraphics[width=0.98\textwidth]{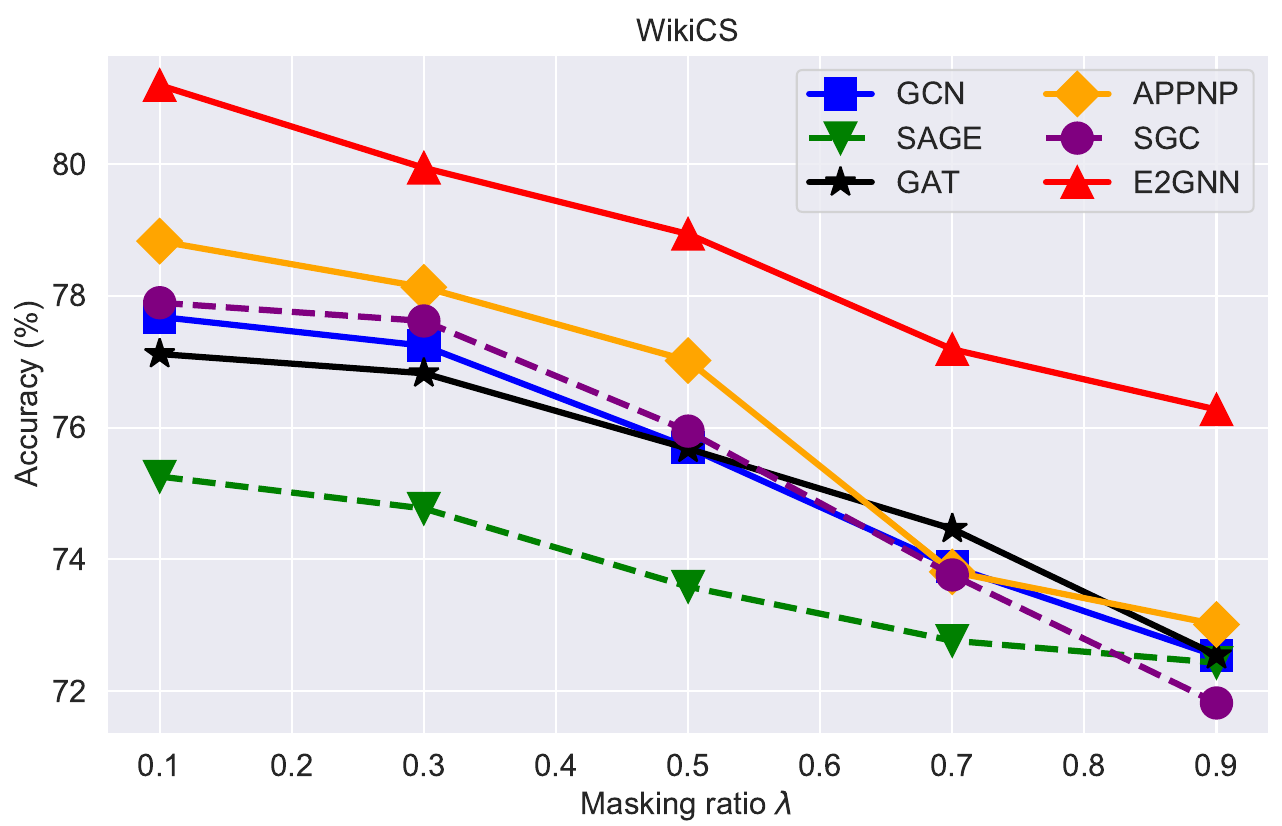}
  \end{subfigure}%
  \caption{Robustness analysis of E2GNN on edge perturbation. }
  \label{fig_robust_sm}
\end{figure}

In addition to the effectiveness of E2GNN, we also conduct a series of experiments to test its robustness in two noisy situations: feature masking and edge perturbation.  For feature masking, we replace $\mathbf{X}$ with $\mathbf{X}\odot\mathbf{M}$, where $\mathbf{M}\in\mathbf{R}^{N\times F}$ is an indicator matrix. Each element in $\mathbf{M}$ is generated by a Bernoulli distribution with parameter $1-\lambda$. For edge perturbation, we randomly mask edges in $\mathcal{G}$ with probability $\lambda$, and take the perturbed version as input. Figure~\ref{fig_robust_fm} and~\ref{fig_robust_sm} report the results of E2GNN on noisy feature and structure, respectively. We make the following observations. 

\textbf{First}, E2GNN is more robust than GNN backbones towards feature noises. In Figure~\ref{fig_robust_fm}, we can see that the performance of all methods tends to decrease when the noise ratio increases. However, E2GNN can still learn from those noisy teachers and achieve better results in general. \textbf{Second}, E2GNN is less sensitive to edge perturbation. As shown in Figure~\ref{fig_robust_sm}, the performance of GNN models drop significantly with the increase of structure noises. Nevertheless, E2GNN performs much better than them, and the performance gap between them is more significant than that in feature noises. These results demonstrate the robustness of E2GNN.

\subsection{Efficiency Analysis (RQ5)}
In this section, we analyze the efficiency of E2GNN compared with standard GNN models and SOTA GNN ensemble efforts. Table~\ref{table_time} lists the inference time for test nodes in the dataset. We can observe that the running time costs of different GNN models are different, and GAT generally runs slower than other methods. Due to the difference, the inference speed of standard GNN ensemble efforts, i.e., GNNE and GEENI, would be heavily affected by the slowest method, so they take significantly more time than individual GNN models. In contrast, by compressing GNN models into unified MLPs, E2GNN can run faster than every GNN teacher. Specifically, E2GNN improves the runtime by 30.1, 15.8, 54.1, 8.0, and 4.0 than GCN, SAGE, GAT, APPNP, and SGC on the Arxiv dataset. It demonstrates the efficiency of E2GNN in assembling GNNs.

\begin{figure}[t]
\centering
\includegraphics[width=7cm]{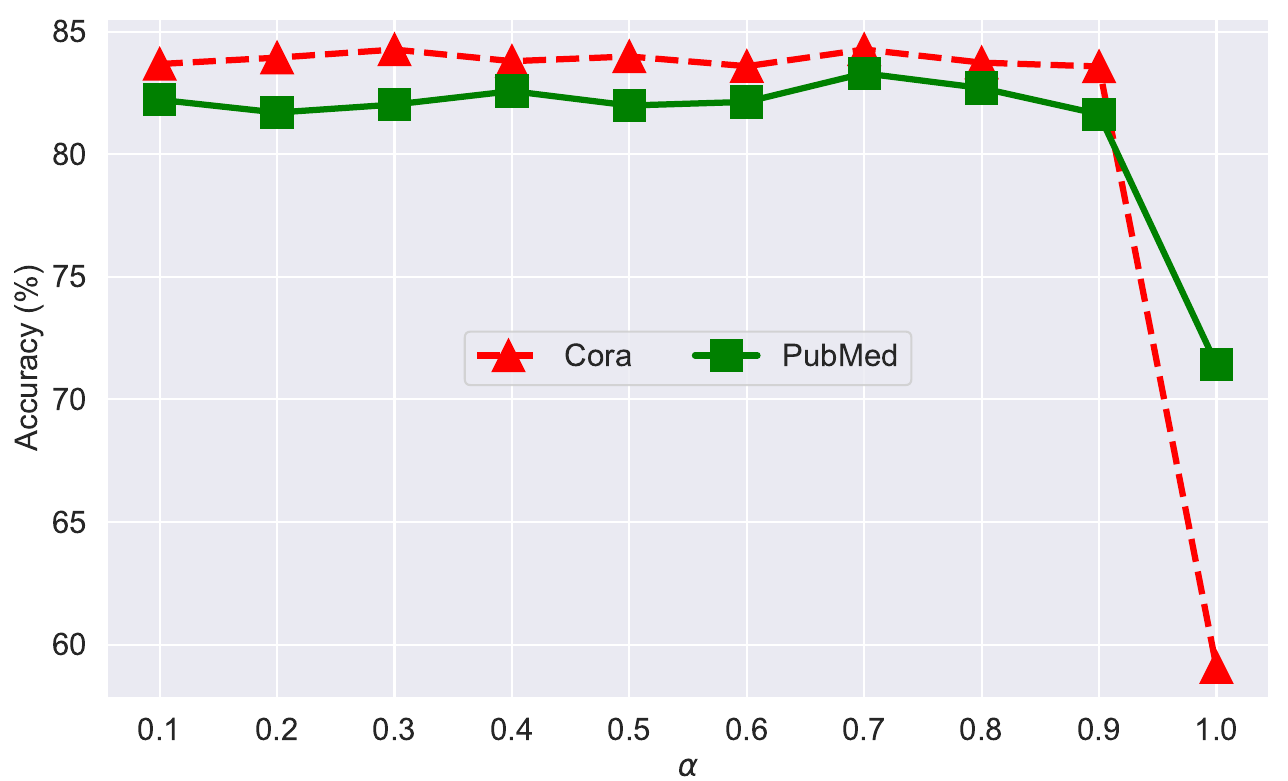}
\caption{The impact of trade-off parameter $\alpha$ on E2GNN. }
\label{fig_parameter}
\end{figure}

\subsection{Further Analysis}
\subsubsection{E2GNN is applicable to other student types} 
We investigate the applicability of E2GNN to another student type, i.e., GNN, and present the results in Table~\ref{table_gnn}. By replacing MLPs with a more advanced student model -- GCN, E2GNN can still achieve better results than GNN teachers. This validates the broader applicability and practicability of the proposed meta-policy. However, we do not observe a significant performance gap between the MLPs and GCN students, which indicates the rationality of compressing GNN models into a unified MLP model.

\subsubsection{E2GNN is robust to the trade-off parameter $\alpha$} $\alpha$ is the trade-off parameter in Eq.~\ref{eq_student}, we investigate its impact on E2GNN and report the results in Figure~\ref{fig_parameter}. We can see that E2GNN generally performs stable when $\alpha$ is from 0.1 to 0.9. When $\alpha=1.0$, E2GNN reduces to vanilla MLPs training based on only labeled nodes. That's why the performance drops substantially. 

\section{Conclusion}
In this work, we improve the effectiveness and efficiency of GNN ensembles under the promising semi-supervised setting with the principle framework of E2GNN. Our method is motivated by the empirical findings that GNN models are complementary with each other in node-level predictions and a large portion of nodes might be simultaneously misclassified by all GNNs. The high-level idea of E2GNN is to selectively distill the knowledge of GNN models in a sample-wise fashion to a lightweight student model. Through extensive experiments across multiple datasets and evaluation settings, we show that E2GNN consistently outperforms state-of-the-art baselines. Our future work would explore extending E2GNN to graph-level learning tasks, e.g., graph classification.

\ifCLASSOPTIONcaptionsoff
  \newpage
\fi



\bibliographystyle{IEEEtran}
\bibliography{sample-base}
\end{document}